\documentclass[10pt,twocolumn,letterpaper]{article}

\usepackage[pagenumbers]{cvpr} 

\usepackage{graphicx}
\usepackage{amsmath}
\usepackage{amssymb}
\usepackage{booktabs}
\usepackage{xcolor}
\usepackage{soul}
\usepackage{comment}

\newcommand\doublecheck{\checked\kern-0.6em\checked}

\setstcolor{red}

\usepackage[pagebackref,breaklinks,colorlinks]{hyperref}

\usepackage[capitalize]{cleveref}
\crefname{section}{Sec.}{Secs.}
\Crefname{section}{Section}{Sections}
\Crefname{table}{Table}{Tables}
\crefname{table}{Tab.}{Tabs.}

\def\cvprPaperID{09235} 
\def\confName{CVPR}
\def\confYear{2023}

\begin{document}

\title{Carpet-bombing patch:\\ attacking a deep network without usual requirements}

\author{Pol Labarbarie\\
IRT SystemX and ONERA/DTIS, Université Paris-Saclay\\
Palaiseau, France\\
{\tt\small firstname.name@irt-systemx.fr}
\and
Adrien Chan-Hon-Tong and Stéphane Herbin\\
ONERA/DTIS, Université Paris-Saclay\\
Palaiseau, France\\
{\tt\small firstname.name@onera.fr}
\and 
Milad Leyli-Abadi\\
IRT SystemX\\
Palaiseau, France\\
{\tt\small firstname.name@irt-systemx.fr}
}
\maketitle

\begin{abstract}
Although deep networks have shown vulnerability to evasion attacks, such attacks have usually unrealistic requirements. Recent literature discussed the possibility to remove or not some of these requirements. This paper contributes to this literature by introducing a carpet-bombing patch attack which has almost no requirement. Targeting the feature representations, this patch attack does not require knowing the network task. This attack decreases accuracy on Imagenet, mAP on Pascal Voc, and IoU on Cityscapes without being aware that the underlying tasks involved classification, detection or semantic segmentation, respectively. Beyond the potential safety issues raised by this attack, the impact of the carpet-bombing attack highlights some interesting property of deep network layer dynamic. 
\end{abstract}


\section{Introduction}\label{sec:intro}
Deep neural networks (DNNs) have given state-of-the-art results in most computer vision tasks, including image classification \cite{krizhevsky2012imagenet}, semantic segmentation \cite{long2015fully}, and object detection \cite{ren2015faster}. Due to their complexity, DNNs have showed vulnerability to adversarial examples, \textit{i.e.}, small perturbations of their inputs designed to fool them \cite{szegedy2013intriguing,biggio2013evasion}. Such vulnerability has motivated researchers to try to develop more robust DNNs \cite{madry2017towards}, as well as to prove that they are robust \cite{cohen2019certified}. Other research has been dedicated to the design of more powerful attacks \cite{kurakin2016adversarial} or the development of different class of attacks, e.g., patch attacks and universal attacks \cite{tramer2020adaptive,brown2017adversarial}. Although such attacks cause safety issues, they also reveal interesting properties about DNNs and their internal structure. For example, invisible noise attacks have highlighted that DNNs sometimes focus on high-frequency \cite{ilyas2019adversarial}. Also, \cite{nguyen2015deep, inkawhich2019feature, lovisotto2022give} have offered connections between adversarial ML and broader topics like network interpretability.

In the same way, we propose in this paper several experiments which highlight that the deep network features are not activated alike by adversarial \textit{clutter} or adversarial \textit{foreground} even when optimised to do so (following \cite{inkawhich2019feature} which introduces the idea to target middle features).


\begin{figure}
    \centering
    \includegraphics[width=\linewidth]{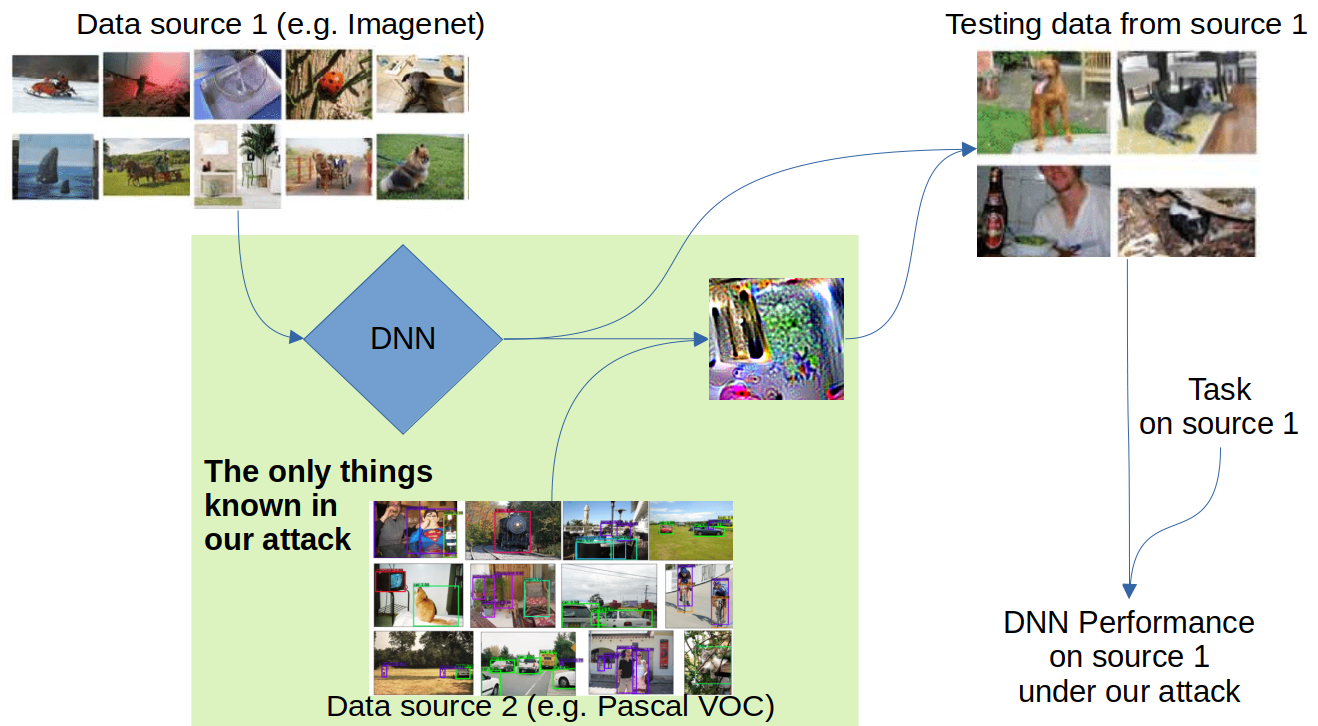}
    \caption{Carpet-bombing patch. Our attack only requires a deep network and a proxy data source to design an adversarial patch. It does not requires neither data from targeted source nor even the knowledge of the task on these data. Yet it could strongly decrease performance on this task.}
    \label{fig:resume}
\end{figure}

\begin{table}
  \centering
    \scalebox{0.70}{
  \begin{tabular}{l|cccc}
    \hline
    Adversarial attacks  & Network & Target image or dataset & Task & Access to pixel  \\\hline
    \cite{szegedy2013intriguing,biggio2013evasion} (original) & +  & +  & + & + \\
    \cite{uesato2018adversarial,ilyas2018black} (blackbox) & $\approx$ & +  & + &  +   \\
    
    \cite{moosavi2017universal} (universal) & + & $\approx$  & + &  +   \\
    \cite{saha2020role,nesti2022evaluating} (patch) & + & $\approx$  & + & \textbf{--} \\
     \cite{rozsa2017lots,inkawhich2019feature} (feature) & \textbf{--} & + & $\approx$ & + \\
    \textbf{Ours} & + & \textbf{--} & \textbf{--} & \textbf{--}\\\hline
  \end{tabular}
  }
  \caption{Typology of adversarial attack depending on the requirement of the attack (less requirement is obviously better). Those requirements are prior knowledge or the fact that the attack is performed on pixel values (not feasible in real world). Our attack is the first with almost not requirement: it is clearly the first tested without knowledge of the task prior (and it also does not require neither the target nor the dataset nor access to pixel).
    }
  \label{tab:recap:attacks}
\end{table}

As major contribution, we show how these experiments led us to find a new kind of adversarial patch attack which requires less prior information than previous attacks as highlighted by Figure \ref{fig:resume} and Table \ref{tab:recap:attacks}. Built to disrupt the feature representations of network encoder, this new patch could affect multiple tasks at the same time. For this reason, this new attack we named carpet-bombing patch, should interest (at least) the safety community. More broadly, it raises questions about intermediary layer dynamics. 

To explain the interest (from a safety point of view) in our patch, we briefly present a typology of the different attacks in the following. The original adversarial attacks present the following requirements:

\begin{itemize}
    \item the knowledge of the targeted testing data,
    \item the knowledge of the targeted task,
    \item the knowledge of the targeted network (architecture, weights and training data),
    \item and, the access to the pixel value (i.e. the attack is performed \textbf{after} image acquisition and not directly in the physical world).
\end{itemize}
Many adversarial attacks try to remove one or more of the above-mentioned requirements. For example, \textit{black box} usually refers to the fact that network is unknown\footnote{Despite it covers situations where queries of the network are allowed, but inner variables of the network are not known, or, the situation where no queries are allowed usually called \textit{transferable attacks}.} \cite{papernot2016transferability}, \textit{universal} usually refers to the fact that the attack is not specific to some target \cite{moosavi2017universal} and \textit{patch-attacks} are more likely to be printable in real-world \cite{brown2017adversarial}.

In this paper, we introduce a new attack that requires almost no prior: this attack does not require knowing the task (to our knowledge, this setting has not been explored), and at the same time, it also does not require knowing the target or to access the pixel. Finally, it offers a moderate effect in blackbox setting. This is the main limitation of this attack regarding recent works like \cite{zhou2018transferable,inkawhich2019feature}. Yet it offers a stronger effect in whitebox setting and is more physically plausible.

The paper is organised as follows: The related works are described in section 2 followed by the scientific story of the proposed attack in section 3. The experiments proving the harmfulness of this attack are presented in section 4. Finally, the conclusions are provided in section 5.

\section{Related works}\label{Related works}

\subsection{Transferable invisible adversarial attacks}

At the beginning, the community tried to build transferable adversarial noises by directly targeting the white box model loss \cite{goodfellow2014explaining,liu2016delving}. Some works improve transferability by adding a momentum \cite{dong2018boosting} or by building an ensemble of white box models on which the attack is built \cite{liu2016delving,xie2019improving}. However, such methods show low transferability due to overfitting of the source model. Zhou et al. \cite{zhou2018transferable} reduce the overfitting by introducing a regulariser that maximises the distance between natural images and adversarial noises in the feature space representation. Rozsa et al. \cite{rozsa2017lots}, and Inkawhich et al. \cite{inkawhich2019feature} propose not to target the white box model loss, but exclusively the feature representations. They propose to minimise the $L_2$ distance between a target point and a source in the feature space for a chosen layer. The source point is often the feature representation of a certain class. This procedure is sensitive to the choice of the target and shows low scalability to larger models, and dataset such as Imagenet \cite{deng2009imagenet}. To better represent the target class in the feature space, Inkawhich et al. \cite{inkawhich2020perturbing} propose to model the class-wise feature distributions of the white box model. Instead of targeting one single layer, they suggest to attack multiple layers. 

\subsection{Adversarial patch attacks (APAs)}

\paragraph{Classification:} APAs were introduced first for image classification by Brown et al. \cite{brown2017adversarial}. Instead of finding a small additive adversarial noise, they constrained the optimization procedure to a small part of the image but allowed it to be unconstrained in magnitude. They produced a patch capable of fooling multiple ImageNet classification models in digital or physical domains (just by printing the patch).

\paragraph{Object detection:} Attacking object detectors was explored in several works working on different applications. In the beginning, patches were directly applied on the struck object. The first two works on patch-based attacks had targeted stop signs \cite{chen2019shape, song2018physical}. They produced stickers, when applied, can fool YOLOv2 or Faster RCNN. Thys et al. \cite{thys2019fooling} were the first to create a patch causing the disappearance of people when it was applied on them. These works focused on designing a patch that overlaps the targeting object to change its class or suppress detection. 

On the other hand, contextual patch attacks are patches which without overlapping with the object of interest can blind the detector. They were first explored by Liu et al. \cite{liu2018dpatch}. Their patch, named Dpatch, showed transferability over patch position, network architecture, and dataset. However, their patches are never clipped to the image range, which is unsuitable for real-world applications. Lee et al. \cite{lee2019physical} studied the Dpatch attack in feasible physical conditions and compared it to their new attack. They outperformed the Dpatch method and showed real-time attack success. The success of those attacks consists of adding a salient patch in the image producing false positives. Saha et al. \cite{saha2020role} were among the first to develop attacks and defence for contextual adversarial patches. They introduce the idea of removing false positives on the patch to measure mainly the contextual effect of patches. 


\paragraph{Semantic segmentation:} The first paper introducing real-world APAs targeting semantic segmentation models is \cite{nesti2022evaluating}. The work presents a novel loss function that, when used, leads to powerful attacks in both digital and real-world scenarios. Unlike patches applied to classification or object detection, the authors showed that semantic segmentation models are not easily corruptible.

As is common throughout the APAs, patches are designed to use the model loss function as the target objective. For comparison, targeting the feature space of the model encoder, we develop a new patch capable of fooling multiple tasks which shows similar performance when attacking with task knowledge. However, we find out that this kind of patches do not keep the model transferability property as is the case with invisible noises.

\section{Another look to feature attacks}

\subsection{Trying to get the best of the two worlds}
\label{subsec:method}
As recalled in related works, the best attacks (from a requirement point of view) were, on one hand, patch attacks \cite{saha2020role}, not requiring to access pixel value (and image target), and, on the other hand, feature-based adversarial attacks \cite{inkawhich2019feature,inkawhich2020perturbing}, which are model transferable. The starting point of our work was to try to combine the best of these two attacks.

More formally, let $F$ a given pre-trained neural network performing a certain task. We denote by $f$ the encoder part of $F$ and denote by $\mathbb{L} = \{ 1, ..., L \}$ the set of the $L$ layers of $f$. One encoder $f$ is often used in multiple $F$, each performing a different task. Let $\delta$ be the adversarial perturbation that can be whether an invisible adversarial noise or an APA. Adversarial noises are obtained by optimising without constraint in space but with a constraint for a certain $L_p$ norm of $\delta$ (i.e $||\delta||_p \leq \varepsilon$ for $\varepsilon>0$). On the other hand, APA is optimised with constraint in space but without constraint in norm. For both of them, to make a realistic attack, we enforce that pixels of the perturbed input or of the APA are in the $[0,1]$ range. Transferable invisible adversarial noises are generally obtained by the following optimisation procedure:

\begin{equation}\label{inkawich_attack}
    \arg\max_{\delta} \mathcal{L}_{task}(F,x,y,\delta) + \eta \mathcal{L}_{feature}(F,x,\delta),
\end{equation}

\noindent where $x$ is the input image, $y$ is the label associated with the task and $\eta > 0$ is the weight corresponding to the contribution of the feature loss. We have two terms: the first one, $\mathcal{L}_{task}$ can be the model loss or a loss derived from the task. This loss disturbs directly the model task. The second term is a feature disruptive loss. This loss generally enforces that the feature map of the perturbed input differs from the original inputs. The adversarial patches are designed with the objective:

\begin{equation}
    \arg\max_{\delta} \mathcal{L}_{task}(F,x,y,\delta).
\end{equation}

In the case of classification, we have $\mathcal{L}_{task} = \mathcal{L}_{cross-entropy}$. Concerning the object detection, the task loss can be general like $\mathcal{L}_{task} = \mathcal{L}_{object-detector}$ or can be more specific such as $\mathcal{L}_{task} = \mathcal{L}_{cross-entropy}$ and for semantic segmentation, we have $\mathcal{L}_{task} \simeq \mathcal{L}_{cross-entropy}$. The transferable invisible attack and the patch attack involve a task loss term that implies knowledge of the task. Our proposed carpet-bombing attack, inspired by the feature disruptive term of transferable noises, is described by the following formula 

\begin{equation}\label{our_attack}
    \arg\max_{\delta} \mathcal{L}_{feature}(f,x,\delta),
\end{equation}

\noindent with

\begin{equation}\label{our_loss}
\mathcal{L}_{feature}(f,x,\delta) = \sum_{l\in \mathbb{L}} \sum_{k\in K} ||(f_l(x_\delta)_k - f_l(x)_k) \odot m_{l}||_2 ,
\end{equation}

\noindent where $f_l(x_\delta)_k$ is the $k$-th feature map of layer $l$ of $f$ for the corrupted image $x_\delta$, $m_l$ the binary mask which is 0 on the patch and 1 everywhere else and $\odot$ is the element-wise product. If $\delta$ is an invisible noise we have 

\begin{equation}
x_\delta = x + \delta 
\end{equation}

\noindent and if, $\delta$ is a patch, we have

\begin{equation}
x_\delta = x \odot (1-m) + \delta \odot m ,
\end{equation}

\noindent where $m$ is a binary mask that is 1 on the patch and 0 everywhere else. Attacking features of $f$ instead of $F$ makes the patch independent of the task considered by $F$. Hence, we can generate one patch capable of fooling multiple $F$ that are based on the same $f$. It can correspond to the scenario where the attacker does not know the task or when multiple tasks use the same encoder $f$.

\subsection{Intriguing behaviour of noise vs patch}
Consequently, following \cite{inkawhich2019feature}, we adapt the idea of feature attack across a patch rather than a noise.
Nevertheless, we observe very different behaviour between these two attacks, although they are basically designed for the same objective. 
We propose in this section to study the effects of the constraint on both the obtained attack and the feature space. We also conduct several experiments to force the patch to have the same effect on features as invisible noise and inversely.

\label{attacking_procedure}
\subsubsection{Attack setting}
Herein, we present the attacking procedure applied to invisible noises and designed patches. Our goal is to design universal contextual attacks. We provide more details in the following concerning these attributes. 

\paragraph{Universal:} We follow \cite{moosavi2017universal,saha2020role} and learn a universal attack (invisible noise or a patch) on the training dataset that works across the unseen test dataset. Instead of finding a specific adversarial $\delta$ for each image $x$, we optimise $\delta$ through iterating over training images. To do so, we sample two subsets of images from the test dataset that the network is designed for: one to train our patch and the other to evaluate it.

\paragraph{Contextual effects:} The objective is to design patches that deteriorate the performance metric by impacting the whole feature map. It should not be sensitive to its placement over an object or the rate of false alarms. For image classification, instead of placing the patch in the middle of the image, we fix the patch at the top-left corner \textit{i.e.} pixel $(5,5)$. For object detection, we extract images wherein objects do not overlap with the patch, which are placed at the top-left corner. Secondly, we remove detections overlapping the patch. Finally, since there are no objects of interest for semantic segmentation, we follow the setup of \cite{nesti2022evaluating}. The patch is centred in the scene, and metrics are not measured on patch pixels.

For both adversarial noises and patches, we solve their corresponding optimisation problem and clip them to $[0,1]$. Clipping ensures that the perturbed image is always maintained in the distribution of original images, and the produced patches are more realistic and do not contain \textit{inf} values. When not specified, the patch is initialised as an all-zeros tensor, and we launch the optimisation process for 100 steps, where 1 step corresponds to 1000 iterations. SGD is used as an optimiser with a momentum of 0.9, a minibatch size of 1, and 10 iterations per minibatch. We evaluate the performance of attacks in the same condition as during the training phase.






\subsubsection{Comparative results}
First, we evaluate the effects of both attacks on the training model and the hidden model (no knowledge of weights or architecture). In our experiments, we use models pre-trained on ImageNet-1K \cite{deng2009imagenet}: ResNet50 (R50) as the whitebox model and ResNet18 (R18) as the hidden model (exactly like in \cite{inkawhich2019feature}). For patches, we use the procedure detailed in Sec. \ref{attacking_procedure} and for invisible noises, following \cite{inkawhich2019feature}, we use iterative gradient sign attack with momentum (TMIFGSM) \cite{dong2018boosting} for 100 steps. Both attacks are targetting uniquely layer 4 of R50 \textit{i.e.}, $\mathbb{L} = \{ L \}$. We split the ImageNet-1K test set into two subsets. We train attacks on 40000 images and evaluate them on 10000 images. Once attacks are learned, we apply them to the testing set. Table \ref{tab:results:noisevspatch} shows that patches have a better severity but present less transferability than invisible noises. 

From this result, we can ask ourselves; \textit{What is making adversarial noises transferable?} 

\begin{table}
  \centering
    \begin{tabular}{|c|c|c|} \hline
     & Whitebox (R50)  & Hidden model (R18)  \\\hline
    Clean  & 76.06   & 70.14  \\\hline
    Noise & 16.13 & \textbf{33.19} \\\hline
    Patch  & \textbf{0.69} & 62.99  \\\hline
  \end{tabular}
  
  \caption{Accuracies (\%) on 10000 ImageNet images for both white box and hidden model under adversarial attacks designed to break internal feature map.}
  \label{tab:results:noisevspatch}
\end{table}

\begin{figure}[t]
  \centering
   \includegraphics[width=1\linewidth]{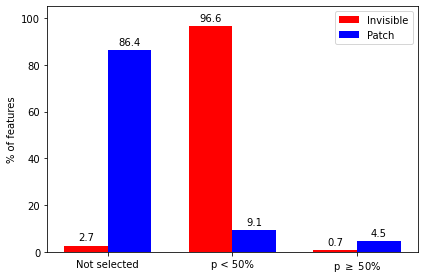}
   \caption{Classification of features for both attacks depending on their frequency of top disruption evaluated on 1000 ImageNet images. The fifty top attacked features for the $L_2$-norm are extracted for each image. The label "Not selected" correspond when the feature does not appear once in the top attacked features through 1000 images.}
   \label{features:repartition}
\end{figure}

\begin{figure*}
  \centering
   \includegraphics[width=1\linewidth]{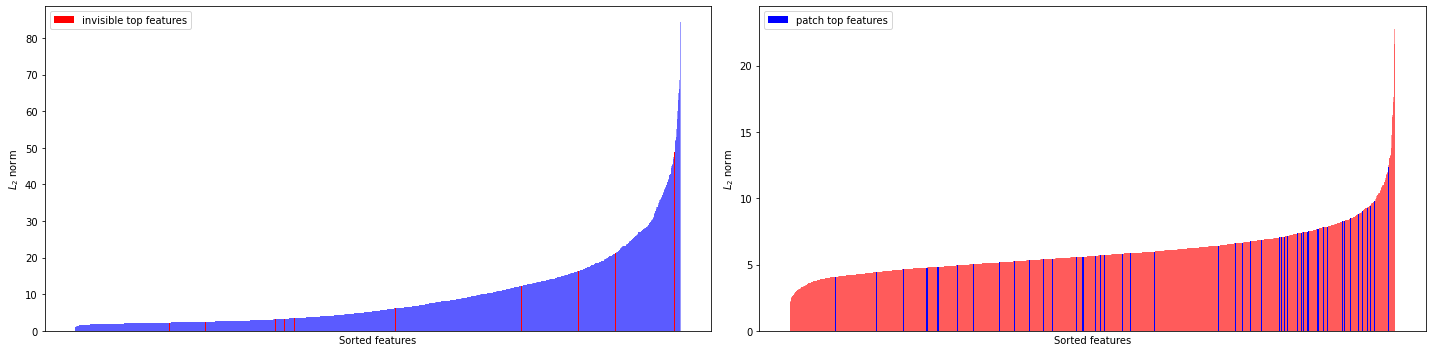}
   \caption{Sorted bar chart of the average $L_2$-distance between cleaned and attacked features over 1000 ImageNet images. On the left, images are corrupted by patch and on the right, by invisible adversarial noise. The top attacked features for invisible noise are highlighted on the left and on the right for the patch.}
   \label{top:features:default}
\end{figure*}

\begin{figure*}
  \centering
   \includegraphics[width=1\linewidth]{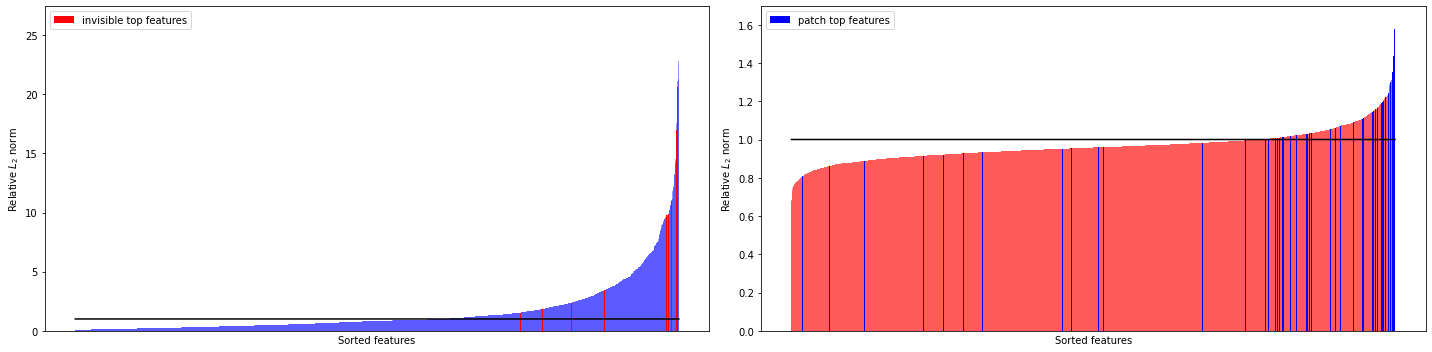}
   \caption{Sorted bar chart of the average relative $L_2$-distance between forced and unforced attack. The average is done on 1000 ImageNet images. On the left, images are corrupted by patch and on the right, by invisible adversarial noise. The top attacked features for invisible noise are highlighted on the left and on the right for the patch. Those top features are used to design the forced attack.}
   \label{top:features:forced}
\end{figure*}

\subsubsection{Attacked features}
To find whether or not attacked features are responsible for the transferability of the attack, we extract "top attacked features" from both types of attacks. To do so, we pass 1000 cleaned and corrupted images to the white box model and extract features from layer 4 of R50. We measure the $L_2$-distance between cleaned and attacked features for each image and save only the top fifty attacked features (\textit{ie} largest norm). We observe from figure \ref{features:repartition} that patches are essentially targeting a limited set of features ($\approx 13 \%$) and focusing on a more significant part of features than invisible noises (4.5 \% against 0.7 \%). The latter targets a more extensive set of features but in a random fashion (96.6 \% for $p < 50\%$). In figure \ref{top:features:default}, we plot the sorted average $L_2$ distance between clean and attacked images for each feature and highlight the top attacked features ($p \geq 50 \%$) in red for the invisible attack and in blue for the APA. We observe that patch attacks significantly impact features more than invisible noise. We retrieve the same dynamic as in figure \ref{features:repartition}: patches focus on a smaller set of features than adversarial noises. By looking at the dynamic of the graph, invisible noises seem to have a diffusive effect on the features and patches show a sharper effect. Interesting to note that they are attacking different features by default.

That brings us to ask: \textit{Are the invisible noises transferable due to the fact that they target these top features?} \textit{Can we force a patch to only target those features? Is this operation make the patch transferable? Are some features insensitive to a particular attack?}

\subsubsection{Mimetic attack} 
We extract the most attacked features for both adversarial attacks ($p \geq 50 \%$), and we denote them, $K_{patch}$ and $K_{inv}$ respectively. Now, we resolve (\ref{our_loss}) the same way as before, except we replace $K$ by $K_{inv}$ when designing the patch and by $K_{patch}$ when designing an invisible attack. Figure \ref{top:features:forced} plots the relative $L_2$ norm when forced attacks target specific features. We highlight the previously top-attacked features for each attack. This graph shows that the patch can attack top invisible features (augmentation from previous trials by a factor of $\simeq 10$ of the attack on those features). It indicates the fact that patches are somehow capable of disrupting top-attacked invisible features. On the other hand, invisible noises could not target other features. Most values around the value one indicate the difficulty of constructing noises targeting a selected set of features. However, does it affect the performance of both models? In Table \ref{tab:results:noisevspatch}, we report performance when we constrain attacks. Both attacks demonstrate less effectiveness on the white box than in default mode. However, concerning the hidden model, we do not observe a gain when targeting invisible features and \textit{vice versa}.

\subsubsection{Spatial impact: } 
At this point, one could wonder if there is a difference between feature perturbation created by noise or patch, as the perturbation designed by an adversarial noise works on different models, which is not the case for an adversarial patch.
One possible explanation is that, the spatial patterns of both perturbations are structurally different (even if this is not visible when considering the spatial average of a given feature channel).

\begin{figure}
  \centering
  \begin{subfigure}{0.48\linewidth}
\includegraphics[width=1\linewidth]{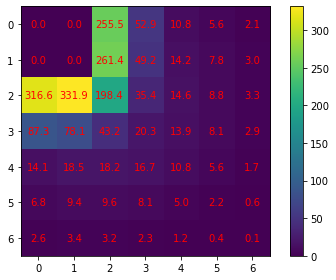}
\caption{Impact map for patch on the whitebox model (R50)}
    \label{fig:map:patch:R50}
  \end{subfigure}
  \hfill
  \begin{subfigure}{0.48\linewidth}
\includegraphics[width=1\linewidth]{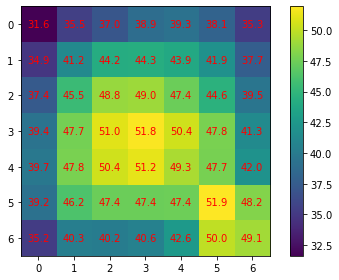}
\caption{Impact map for noise on the whitebox model (R50)}
    \label{fig:map:invisible:R50}
  \end{subfigure}
  
    \begin{subfigure}{0.48\linewidth}
\includegraphics[width=1\linewidth]{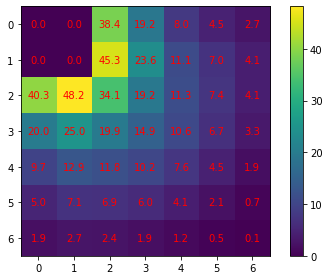}
\caption{Impact map for patch on the hidden model (R18)}
    \label{fig:map:patch:R18}
  \end{subfigure}
  \hfill
  \begin{subfigure}{0.48\linewidth}
\includegraphics[width=1\linewidth]{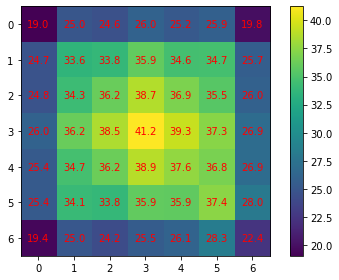}
\caption{Impact map for noise on the hidden model (R18)}
    \label{fig:map:invisible:R18}
  \end{subfigure}
  \caption{Impact map obtained by averaging the $L_2$-distance over features in cells. On the top row are represented maps for the different attacking procedures (patch or invisible) for the whitebox model (R50) and on the bottom row for the hidden model (R18).}
  \label{fig:maps}
\end{figure}

We propose to study the spatial impact of attacks. We apply attacks built by default (\textit{i.e.,} no constraint on the attacked features) on 1000 images and measure the average $L_2$-distance between attacked features and clean features in each cell of the feature map. We obtain a map representing the spatial impact of attacks (Fig. \ref{fig:maps}). The heat maps of attack impact are shown in top figures when applied to the white box model (Fig. \ref{fig:map:patch:R50}, \ref{fig:map:invisible:R50}) and on the bottom, the relative map  when applied on the hidden and the white box model (Fig. \ref{fig:map:patch:R18}, \ref{fig:map:invisible:R18}). From these plots (Fig. \ref{fig:map:patch:R50}, \ref{fig:map:invisible:R50}), it could be clearly seen that patches are attacking their neighbourhood area and the adversarial noises are diffused over all features. We see that patches have a broader impact on attacked cells than noises. But is this significant impact capable of transfer to the hidden model? From plots \ref{fig:map:patch:R18} and \ref{fig:map:invisible:R18}, we observe that the patch effect on its neighbourhood almost disappears. Contrarily, concerning the invisible noises, their effect decreases despite their significant impact. 


\subsection{Transfer between related networks}

Seeing the behaviour of the patch attack on a hidden model, we consider different but closer models.
For this purpose, we consider a Darknet-19 \cite{redmon2017yolo9000} classifier trained on Imagenet, and we finetune it into a Yolo on PASCAL VOC. We save a snapshot of the Darknet-19 being finetuned every 300 iterations and measure the impact of the original patch designed to target the Imagenet.
Precisely, we choose to target last layer \textit{i.e.,} $\mathbb{L} = \{L \}$). We pass 1000 ImageNet images and extract the top attacked features. 

\begin{figure*}
  \centering
  \begin{subfigure}{0.33\linewidth}
\includegraphics[width=1\linewidth]{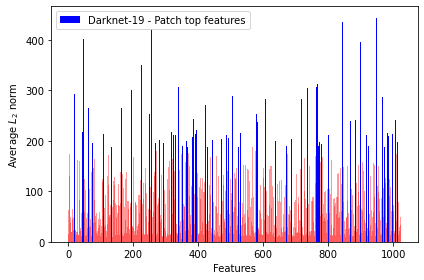}
\caption{Darknet-19}
    \label{fig:feature:darknet}
  \end{subfigure}
  \hfill
  \begin{subfigure}{0.33\linewidth}
\includegraphics[width=1\linewidth]{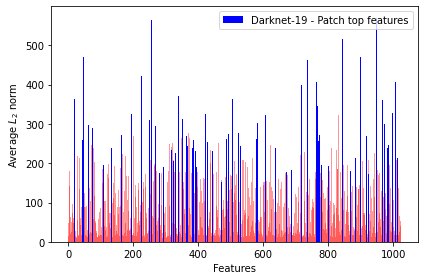}
\caption{YoloV2 trained on 300 batches}
    \label{fig:feature:yolo300}
  \end{subfigure}
  \hfill
  \begin{subfigure}{0.33\linewidth}
\includegraphics[width=1\linewidth]{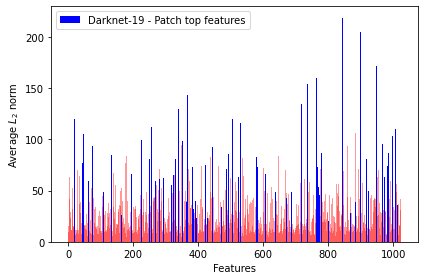}
\caption{YoloV2 trained on 600 batches}
    \label{fig:feature:yolo600}
  \end{subfigure}
  
    \begin{subfigure}{0.33\linewidth}
\includegraphics[width=1\linewidth]{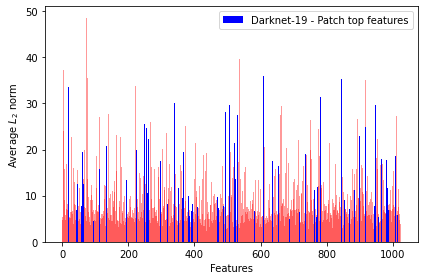}
\caption{YoloV2 trained on 900 batches}
    \label{fig:feature:yolo900}
  \end{subfigure}
  \hfill
  \begin{subfigure}{0.33\linewidth}
\includegraphics[width=1\linewidth]{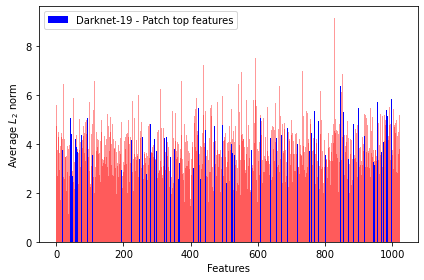}
\caption{YoloV2 trained on 30000 batches}
    \label{fig:feature:yolo30000}
  \end{subfigure}
  \hfill
  \begin{subfigure}{0.33\linewidth}
\includegraphics[width=1\linewidth]{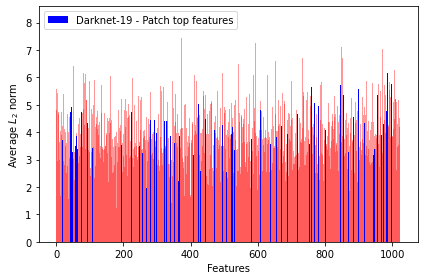}
\caption{YoloV2 final version}
    \label{fig:feature:yolo}
  \end{subfigure}
  \caption{Patch impact evolution during the fine-tuning of Darknet-19 to YoloV2, from top-left to bottom-right. For each step, a bar chart of the average $L_2$-distance between cleaned and attacked features for the last layer of Darknet-19. The average is computed over 1000 images. Highlighted the top attacked features by patch attack designed to disrupt Darknet-19 last layer features.}
  \label{fig:feature:darknet:to:yolo}
\end{figure*}

As a result, figure \ref{fig:feature:darknet:to:yolo} shows the evolution of patch feature impact for Darknet-19 and evolutionary versions of YoloV2. This graph shows that patch impact decreases very quickly during YoloV2 training. First, the amplitude of the attack strongly decreases. Then, we observe that the patch has a significant pattern drift considering most attacked features after 600 iterations. So, even when targeting the same model with related weights, the original patch quickly becomes ineffective (we check that it is possible to design a specific patch for each snapshot). 

These experiments also point out that the features not considered by the attack in the original model can become the most important. At this point, it means either weight changes too quickly so that the initial relationship between the two networks is irrelevant or that the network importantly reorganises the influence of features.


\subsection{Carpet-bombing patch}
Seeing previous experiments, it seems that combining the best of the two worlds does not provide satisfying results: the conversion proposed in \cite{inkawhich2019feature} into a patch attack (to remove additional requirements) does not inherit from \cite{inkawhich2019feature,inkawhich2020perturbing} property. Nevertheless, even though our patch attack provides only a moderate effect in black box setting, a second look at the noise-vs-patch experiments reveals some interesting properties.

First, our patch attack is much more powerful than noise based for modifying the feature map in white box setting: in this setting, the feature norm modification is eight times more intense with the patch than noise (in absolute value).
This is why we call our patch a carpet bombing patch: it heavily modifies the targeted feature maps, eventually producing an output modification. Moreover, we observe that this heavy perturbation offers some interesting features:
\begin{itemize}
    \item like most patch attacks, it does not need to know the target image, but even stronger, it can be designed from a proxy dataset;
    \item it does not need to know the underlying task;
    \item and as a patch attack, it does not need to access the pixel.
\end{itemize}
To our knowledge, those three features have never been observed simultaneously: standard attacks rely on the task loss and even \cite{inkawhich2019feature} which targets features only consider classification, and crafting a patch on proxy data source has never been explored before (of course, the two data sources can not be too different).

Despite combining network transferable attack \cite{inkawhich2019feature} and patch attack \cite{saha2020role} was not satisfying, we still obtained a threatening attack (described in \ref{subsec:method}) whose performances are detailed in the next section.

\section{Experiments}
In the previous section, we have shown a first contribution\footnote{potentially interesting for all computer vision community} by comparing the behaviour of \textit{adversarial noise} and \textit{adversarial shape}. Yet, the main contribution, oriented toward the safety community, is the design of a new adversarial patch attack with even fewer requirements than previous ones (as pointed in table \ref{tab:recap:attacks}). Numerous experiments are performed in this section highlighting that the proposed attack requires neither the underlying task, the target, or even the knowledge of the exact data source.

\subsection{Datasets} We report our results for image classification task on the commonly-used dataset ImageNet \cite{deng2009imagenet}, for object detection task on PASCAL VOC \cite{everingham2008pascal} and for semantic segmentation task on Cityscapes \cite{cordts2016cityscapes} which is a popular dataset for urban semantic segmentation. 

Briefly, ImageNet is a set of 1M high resolution images (256x256 pixels) tagged with 1000 labels. Pascal VOC is a set of 50K large images (some above 512x512 pixels) containing a few objects from 10 classes. Finally, Cityscapes consists of a few thousand high-resolution images ($1024 \times 2048$) taken from a car while driving (there are 2975 images for training and 500 for validation).

\subsection{Eliminating the requirement of knowing the task}

In this section, our goal is to design a patch capable of fooling multiple tasks without any underlying knowledge of them. In the following, we explain each task and the corresponding results.  

\subsubsection{Image classification} We design our patch to attack the backbone of the well-known ResNet-50 \cite{he2016deep} from Pytorch Model Zoo. This model has been pretrained on ImageNet-1K \cite{deng2009imagenet}. We split the ImageNet-1K test set into training and test sets. The patch is fixed at the top-left corner i.e., pixel $(5,5)$ with a dimension $50 \times 50$. We solve Equation (\ref{our_loss}) using the previously explained procedure (see Sec. \ref{attacking_procedure}) and choose to target only layer 4. We compare our attack to the well-known patch attack for classification \cite{brown2017adversarial}. 

For this task, the model accuracy has dropped nearly to 0\% (Tab \ref{tab:results:multi:tasks}). This result is impressive since our patch has been designed without any knowledge of the underlying task. It can deteriorate the feature map so the network can not exploit it.

\subsubsection{Object detection} Following the methodology and the conditions used in \cite{saha2020role}, we compare our attack against their universal blindness attack, where both methods target YoloV2 architecture \cite{redmon2017yolo9000}. We sample from the PASCAL VOC \cite{everingham2008pascal} test dataset, two subsets of images that do not overlap with the patch. Each image is rescaled to $416 \times 416$ dimensions, and we fixed each patch to $100 \times 100$ dimensions at the top-left corner. Again, we solve Equation (\ref{our_loss}) using the previously mentioned procedure (Sec. \ref{attacking_procedure}). For the encoder part ($f$) of YoloV2 architecture ($F$), Darknet-19 is considered. We target the last layer of Darknet-19, i.e., $\mathbb{L} = \{L\}$. Once the patches are designed and learned by optimising the Equation (\ref{our_loss}) and \cite{saha2020role}, we evaluate them with the same emplacement as during the training phase.
In evaluation setting, we set the confidence threshold of YoloV2 architecture to $0.0005$, the Non-Maximum Suppression (NMS) to 0.45, and the Intersection Over Union (IOU) to $0.5$.

The corresponding results are shown in Table \ref {tab:results:multi:tasks}. Our attacking performance is interesting because this performance is reached without introducing false alarms to the patch. The obtained result proves the weakness of object detectors, which indicates that the disruption of feature representations can influence the decision of a complicated task like detection. 

\begin{table}
  \centering
     \scalebox{0.66}{
  \begin{tabular}{|l|cc|cc|cc|}
    \toprule
     Task &  \multicolumn{2}{|c|}{Clean} &  \multicolumn{2}{|c|}{SOTA attack}  & \multicolumn{2}{|c|}{Ours}  \\
    \midrule
    Classification (Acc)   & \multicolumn{2}{|c|}{76.06}   & \multicolumn{2}{|c|}{0.17 \cite{brown2017adversarial}} & \multicolumn{2}{|c|}{0.69} \\
    \midrule
    Detection (mAP) & \multicolumn{2}{|c|}{72.77}  & \multicolumn{2}{|c|}{52.29 \cite{saha2020role}} & \multicolumn{2}{|c|}{59.04}   \\
    \midrule

    Segmentation (mIOU/mAcc)  & 69.00 & 78.00 & 44.59 & 54.54 \cite{nesti2022evaluating}  & 43.11 & 54.81   \\
    \bottomrule
  \end{tabular}}
  \caption{Comparison of performance (\%) under our attack and state-of-the-art task patch attacks for image classication, object detection and semantic segmentation. SOTA is different for each task, while our attack is unaware of the underlying task.}
  \label{tab:results:multi:tasks}
\end{table}

\subsubsection{Semantic segmentation} 
We compare the performance of the proposed patch against the recent state-of-the-art patch attacks designed for semantic segmentation task \cite{nesti2022evaluating} and using Cityscapes dataset\cite{cordts2016cityscapes}. We use the same settings as in \cite{nesti2022evaluating} to compare. For patch training, we randomly sample 250 images from the training set and to evaluate the impact of patches, we use the entire validation set. For the sake of comparison, we select BiSeNet \cite{yu2018bisenet}, one of the state-of-the-art real-time semantic segmentation models. We target the two last layers of the Context Path module, i.e., $\mathbb{L} = \{L-1, L\}$. Since we are not using Expectation Over Transformation (EOT) \cite{athalye2018synthesizing}, patches are placed at the middle part of images following \cite{nesti2022evaluating} and have a dimensionality of $300 \times 600$ pixels. We use Adam optimiser with a learning rate of $0.5$ and run the optimisation process over 200 epochs. The evaluation has been done with the same emplacement of image as during the training phase.

Table \ref{tab:results:multi:tasks} shows that we obtain similar results to state-of-the-art segmentation attacks. Disrupting features of one module of the model seems to degrade the performance highly.

\subsection{Removing the requirement of knowing the data distribution}
Finally, we test whether or not our attack could rely on data from the targeted distribution. In a real-life scenario, a hacker would often have access to the underlying target model than to the data on which the model was built. We evaluate the impact of our attack when trained on a completely different dataset. We considered two tasks and three datasets. For image classification, our patch is built on PASCALVOC \cite{everingham2008pascal}, or on Cityscapes \cite{cordts2016cityscapes} to sway R50 trained on ImageNet \cite{deng2009imagenet}. To target YoloV2 architecture \cite{redmon2017yolo9000} trained on PASCALVOC \cite{everingham2008pascal}, we design our attack on ImageNet \cite{deng2009imagenet} or on Cityscapes \cite{cordts2016cityscapes}. We used the optimisation procedure described in section \ref{attacking_procedure}. Once patches are learned, we apply them to the data on which the model is trained. For clarity, we report the performance of models when attacked by patches designed on the targeted distribution.  

We show impressive attacking results for both tasks (Tab \ref{tab:results:dataset:transfer}). Our patch decreases the performance near to 0\%, and by 9 \% points for classification and detection, respectively. We demonstrate similar results for classification when the patch is directly learned on the targeted distribution. And for detection, mean average precision (mAP) falls significantly independently from the fact that the patch is designed on ImageNet, Cityscapes or PASCALVOC. 

To the best of our knowledge, such a level of degradation without knowledge of the task, the target (or the dataset of the target) and without direct access to the image pixel has never been reported before.

\begin{table}
  \centering
    \scalebox{0.58}{

    \begin{tabular}{|c|c|c|c|c|} \hline
   Task  & Clean &ImageNet $\xrightarrow[]{} \mathcal{D}$  & PASCALVOC $\xrightarrow[]{} \mathcal{D}$  & Cityscapes $\xrightarrow[]{} \mathcal{D}$  \\\hline
    Classification (Acc) - R50  & 76.06 & 0.69   & 0.26 &  0.48  \\\hline
    Detection (mAP) - YoloV2 & 72.77 & 64.33 & 59.04 & 63.10 \\\hline
  \end{tabular}
  }
  \caption{Comparison of performance (\%) under our attack when the targeted dataset is not known. The targeted dataset $\mathcal{D}$ is ImageNet and PASCALVOC for classification and detection, respectively.}
  \label{tab:results:dataset:transfer}
\end{table}

\section{Conclusion}
This paper introduces a new evasion attack targeting deep networks, which can be crafted without access to targeted datum (or data distribution), targeted tasks and that could plausibly be produced in the real world. Such easily reproducible attack should be considered for safety reasons. Beyond the proposed attack, this paper also reports interesting experiments which highlight the difference between adversarial noise and adversarial patch. These results may interest more broadly than the strict attack-defence game and should be deeply studied in future works.

\section*{Acknowledgements}
This work has been supported by the French government under the France 2030 program, as part of the SystemX Technological Research Institute. 

{\small
\bibliographystyle{ieee_fullname}
\bibliography{egbib}
}

\end{document}